\def\year{2020}\relax
\documentclass[letterpaper]{article} % DO NOT CHANGE THIS
\usepackage{aaai20}  % DO NOT CHANGE THIS
\usepackage{times}  % DO NOT CHANGE THIS
\usepackage{helvet} % DO NOT CHANGE THIS
\usepackage{courier}  % DO NOT CHANGE THIS
\usepackage[hyphens]{url}  % DO NOT CHANGE THIS
\usepackage{graphicx} % DO NOT CHANGE THIS
\usepackage{graphicx}  % DO NOT CHANGE THIS
\usepackage{multirow}

\usepackage{amsfonts}
\usepackage{amsmath}
\usepackage{algorithm}
\usepackage[noend]{algorithmic}

\title{ICD Coding from Clinical Text Using Multi-Filter \\Residual Convolutional Neural Network}
\author{
\Large \textbf{Fei Li,\textsuperscript{\rm 1} Hong Yu\textsuperscript{\rm 1,2,3,4}}\\ % All authors must be in the same font size and format. Use \Large and \textbf to achieve this result when breaking a line
\textsuperscript{\rm 1}Department of Computer Science, University of Massachusetts Lowell, Lowell, MA, United States\\
\textsuperscript{\rm 2}Center for Healthcare Organization and Implementation Research, Bedford Veterans Affairs Medical Center, Bedford, MA, United States\\
\textsuperscript{\rm 3}Department of Medicine, University of Massachusetts Medical School, Worcester, MA, United States\\
\textsuperscript{\rm 4}School of Computer Science, University of Massachusetts, Amherst, MA, United States\\%If you have multiple authors and multiple affiliations
fei\_li, hong\_yu@uml.edu % email address must be in roman text type, not monospace or sans serif
}
 
\begin{document}

\maketitle

\begin{abstract}
Automated ICD coding, which assigns the International Classification of Disease codes to patient visits, has attracted much research attention since it can save time and labor for billing. The previous state-of-the-art model utilized one convolutional layer to build document representations for predicting ICD codes. However, the lengths and grammar of text fragments, which are closely related to ICD coding, vary a lot in different documents. Therefore, a flat and fixed-length convolutional architecture may not be capable of learning good document representations. In this paper, we proposed a \textbf{Multi}-Filter \textbf{Res}idual \textbf{C}onvolutional \textbf{N}eural \textbf{N}etwork (MultiResCNN) for ICD coding. The innovations of our model are two-folds: it utilizes a multi-filter convolutional layer to capture various text patterns with different lengths and a residual convolutional layer to enlarge the receptive field. We evaluated the effectiveness of our model on the widely-used MIMIC dataset. On the full code set of MIMIC-III, our model outperformed the state-of-the-art model in 4 out of 6 evaluation metrics. On the top-50 code set of MIMIC-III and the full code set of MIMIC-II, our model outperformed all the existing and state-of-the-art models in all evaluation metrics. The code is available at https://github.com/foxlf823/Multi-Filter-Residual-Convolutional-Neural-Network.

\end{abstract}

\section{Introduction}

The International Classification of Diseases (ICD), which is organized by the World Health Organization, is a common coding method used in various healthcare systems such as hospitals. It includes many pre-defined ICD codes which can be assigned to patients' files such as electronic health records (EHRs). These codes represent diagnostic and procedural information during patient visits. Healthcare providers and insurance companies need these information to diagnose patients and bill for services \cite{bottle2008intelligent}. However, manual ICD coding has been demonstrated to be labor-consuming and costly \cite{o2005measuring}.

The research community has investigated a number of approaches for automated ICD coding, including the models based on both traditional machine learning \cite{perotte2013diagnosis,kavuluru2015empirical} and deep learning \cite{shi2017towards,xie2018neural}. In terms of data, prior work utilized different domains of data such as radiology reports \cite{pestian2007shared} and death certificates \cite{koopman2015automatic}, and different modal data such as structured \cite{perotte2013diagnosis} and unstructured text \cite{scheurwegs2017selecting}. Moreover, some previous work adopted full ICD codes to perform this task \cite{baumel2018multi} while other work adopted partial codes \cite{xu2018multimodal}. Due to such situation, it is difficult to directly compare different work. In this paper, we followed the line of predicting ICD codes from unstructured text of the MIMIC dataset \cite{johnson2016mimic}, because it is widely studied and publicly available.

\begin{table}[t]
\caption{Examples of clinical text fragments and their corresponding ICD codes.}\smallskip
\centering
\resizebox{.95\columnwidth}{!}{
\smallskip
\begin{tabular}{l}
\hline
\emph{998.32: Disruption of external operation wound}\\
... wound infection, and \textbf{wound breakdown} ...\\
\hline
\emph{428.0: Congestive heart failure}\\
... DIAGNOSES: 1.  \textbf{Acute congestive heart failure}\\
2.  Diabetes mellitus 3.  Pulmonary edema ...\\
\hline
\emph{202.8: Other malignant lymphomas} \\
... a 55 year-old female with \textbf{non Hodgkin's lymphoma}\\
and acquired C1 esterase inhibitor deficiency ...\\
\hline
\emph{770.6: Transitory tachypnea of newborn} \\
... Chest x-ray was consistent with \textbf{transient tachypnea}\\
\textbf{of the newborn} ...\\
\hline
\emph{424.1: Aortic valve disorders}\\
... mild \textbf{aortic stenosis with an aortic valve area} of\\
1.9 cm squared and 2+ \textbf{aortic insufficiency} ...\\
\hline
\end{tabular}
}
\label{table:example}
\end{table}

The state-of-the-art model for this line of work is the combination of the convolutional neural network (CNN) and the attention mechanism \cite{mullenbach2018explainable}. However, this model only contains one convolutional layer to build document representations for subsequent layers to predict ICD codes. As shown in Table \ref{table:example}, ICD-related text spans and patterns vary in different examples. Therefore, it may not be sufficient to learn decent document representations from a flat and fixed-length convolutional architecture.

In this paper, we proposed a \textbf{Multi}-Filter \textbf{Res}idual \textbf{C}onvolutional \textbf{N}eural \textbf{N}etwork (MultiResCNN) for ICD coding using clinical discharge summaries. Our MultiResCNN model is composed of five layers: the input layer leverages word embeddings pre-trained by word2vec \cite{mikolov2013distributed}; the multi-filter convolutional layer consists of multiple convolutional filters \cite{kim2014convolutional}; the residual convolutional layer contains multiple residual blocks \cite{he2016deep}; the attention layer keeps the interpretability for the model following \cite{mullenbach2018explainable}; the output layer utilizes the sigmoid function to predict the probability of each ICD code.

Our main contribution is that we proposed a novel CNN architecture that combines the multi-filter CNN \cite{kim2014convolutional} and residual CNN \cite{he2016deep}. The advantages are two-folds: MultiResCNN not only captures various text patterns with different lengths via the multi-filter CNN, but also enlarges the receptive field\footnote{http://cs231n.github.io/convolutional-networks/} \cite{garcia2004convolutional} via the residual CNN. Thus, our model can benefit from rich patterns, the large receptive field and deep architecture. Such method has achieved great success in natural language processing \cite{vaswani2017attention} and computer vision \cite{NIPS2012_4824}.

To evaluate our model, we employed the MIMIC dataset \cite{johnson2016mimic} which has been widely used for automated ICD coding. Compared with 5 existing and state-of-the-art models \cite{perotte2013diagnosis,prakash2017condensed,shi2017towards,baumel2018multi,mullenbach2018explainable}, our model outperformed them in nearly all the evaluation metrics (i.e., macro- and micro-AUC, macro- and micro-F1, precision at K). Concretely, in the MIMIC-III experiment using full codes, our model outperformed these models in macro-AUC, micro-F1 and precision at 8 and 15. In the MIMIC-III experiment using top-50 codes and the MIMIC-II experiment using full codes, our model outperformed these models in all evaluation metrics. Moreover, hyper-parameter tuning experiments show that the multi-filter and residual convolutional layers help our model to improve its performance significantly.

\section{Related Work}

To the best of our knowledge, the earliest work of automated ICD coding was proposed by Larkey and Croft \shortcite{larkey1996combining}. They combined three classifiers, K-nearest-neighbor, relevance feedback and Bayesian independence, to assign ICD9 codes to inpatient discharge summaries. However, their method only assigns one code to each discharge summary. Pestian et al. \shortcite{pestian2007shared} organized a shared task of assigning ICD-9 codes to radiology reports and their task requires models to assign a large set of codes to each report.

Early work usually used supervised machine learning approaches for ICD coding. Perotte et al. \shortcite{perotte2013diagnosis} leveraged ``flat'' and ``hierarchical'' Support Vector Machines (SVMs) for automatically assigning ICD9 codes to the discharge summaries of the MIMIC-II repository \cite{johnson2016mimic}. Their results show that the hierarchical SVM performs better than the flat one. Kavuluru et al. \shortcite{kavuluru2015empirical} used the unstructured text in 71,463 EMRs, which come from the University of Kentucky Medical Center, to evaluate supervised learning approaches such as multi-label classification and learning to rank for the ICD9 code assignment. Koopman et al. \shortcite{koopman2015automatic} employed the SVM to identify cancer-related causes of death
from 447,336 death certificates. Their model is cascaded: the first one identified the presence of cancer and the second identified the type of cancer according to the ICD-10 classification system. Scheurwegs et al. \shortcite{scheurwegs2017selecting} evaluated coverage-based feature selection methods and Random
Forests on seven medical specialties for ICD9 code prediction and two for ICD10, incorporating structured and unstructured text.

With the development of deep learning, researchers also explored neural networks for this task. Shi et al. \shortcite{shi2017towards} utilized the long short-term memory (LSTM) and attention mechanism for automated ICD coding from diagnosis descriptions. Xie and Xing \shortcite{xie2018neural} also adopted the LSTM but they introduced the tree structure and adversarial learning to utilize code descriptions. Prakash et al. \shortcite{prakash2017condensed} exploited condensed memory neural networks and evaluated it on the free-text medical notes of the MIMIC-III dataset. Baumel et al. \shortcite{baumel2018multi} proposed a hierarchical gated recurrent unit (GRU) network, which encodes sentences and documents with two stacked layers, to assign multiple ICD codes to discharge summaries of the MIMIC II and III datasets. Mullenbach et al. \shortcite{mullenbach2018explainable} incorporated the convolutional neural network (CNN) with per-label attention mechanism. Their model achieved the state-of-the-art performance among the work using only unstructured text of the MIMIC dataset. Xu et al. \shortcite{xu2018multimodal} built a hybrid system that includes the CNN, LSTM and decision tree to predict ICD codes from unstructured, semi-structured and structured tabular data. In addition, Lipton et al. \shortcite{lipton2015learning} utilized LSTMs to predict diagnostic codes from time series of clinical measurements, while our work focuses on text data.

\section{Method}
\begin{figure}[t]
\centering
\includegraphics[width=.95\columnwidth]{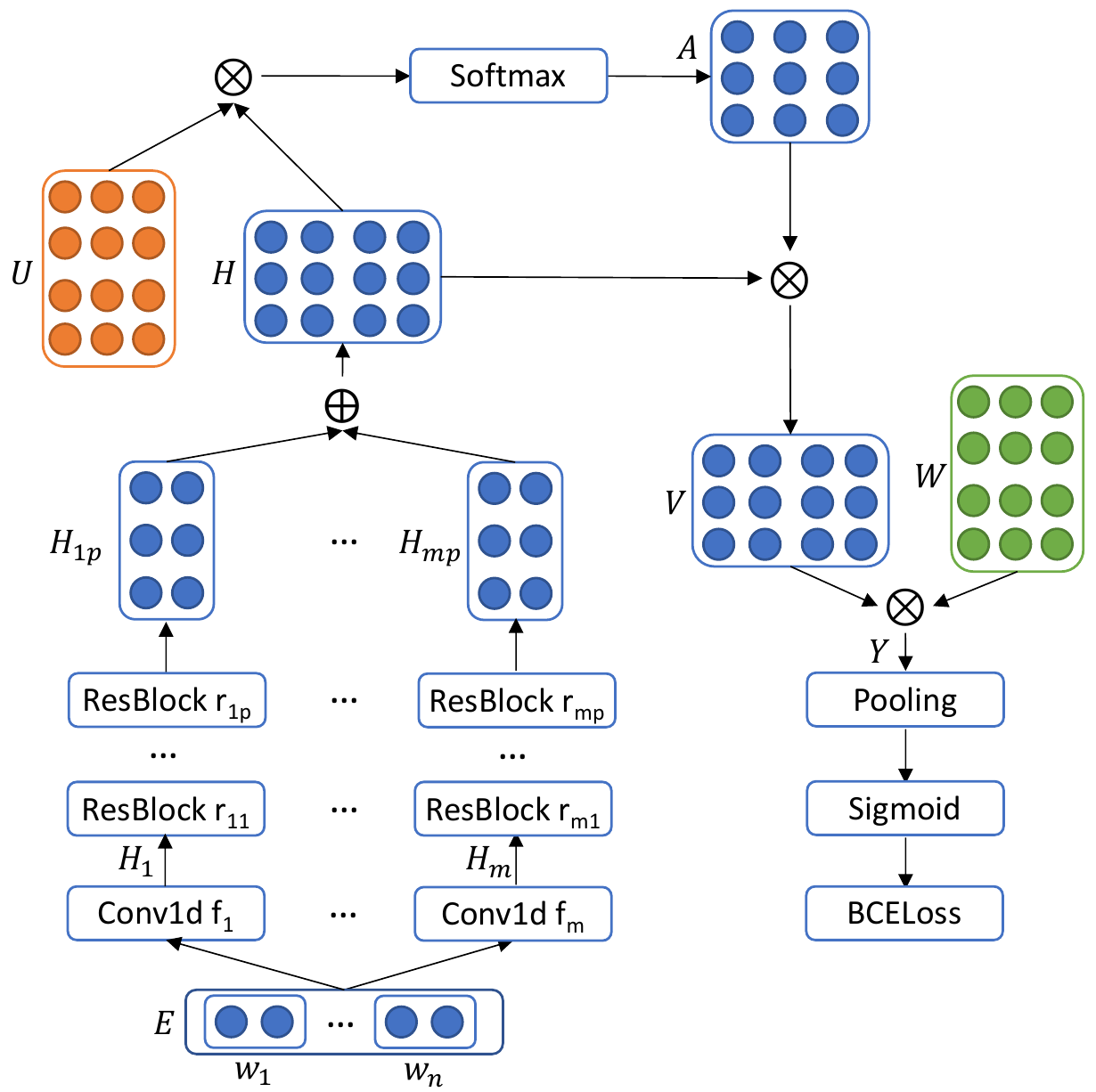} 
\caption{The architecture of our MultiResCNN model. ``Conv1d'' represents the 1-dimensional convolution, ``ResBlock'' represents the residual block, ``$\oplus$'' represents the concatenation operation and ``$\otimes$'' represents the matrix multiplication. Here we use orange and green for $U$ and $W$ to denote they are learnable parameters, and to distinguish with other matrices (e.g., $H$) which are not parameters. 
}
\label{fig:model}
\end{figure}

In this section, we will introduce our \textbf{Multi}-filter \textbf{Res}idual \textbf{C}onvolutional \textbf{N}eural \textbf{N}etwork (MultiResCNN), whose architecture is shown in Figure \ref{fig:model}. Throughout this paper, we employed the following notation rules: matrices are written as italic uppercase letters (e.g., $X$); vectors and scalars are written as italic lowercase letters (e.g., $x$).

\subsection{Input Layer}
Our model leverages a word sequence $w=\{w_1, w_2, ..., w_n\}$ as input, where n denotes the sequence length. Assuming that $\tilde{E}$ denotes the word embedding matrix, which is pre-trained via word2vec \cite{mikolov2013distributed} from the raw text of the dataset. A word $w_n$ will correspond to a vector $e_n$ by looking up $\tilde{E}$. Therefore, the input will be a matrix $E=\{e_1, e_2, ..., e_n\} \in \mathbb{R}^{n\times d^{e}}$.

\subsection{Multi-Filter Convolutional Layer}

\begin{figure}[t]
\centering
\includegraphics[width=.95\columnwidth]{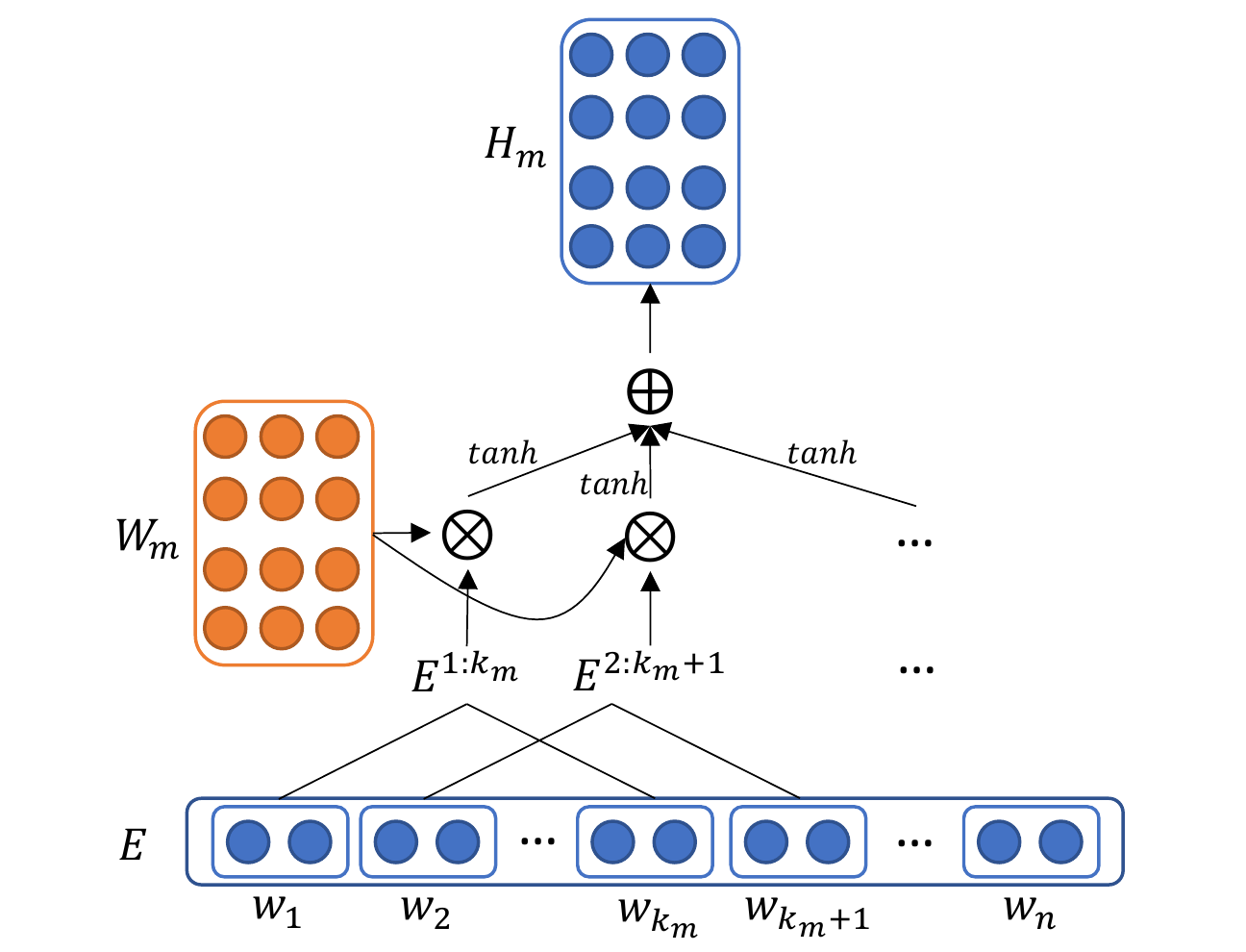} 
\caption{The architecture of a 1-dimensional convolution filter $f_m$. ``$\oplus$'' represents the concatenation operation and ``$\otimes$'' represents the matrix multiplication.}
\label{fig:conv}
\end{figure}

To capture the patterns with different lengths, we leveraged the multi-filter convolutional neural network \cite{kim2014convolutional}, where each filter has a different kernel size (i.e., word window size). Assuming we have $m$ filters $f_1, f_2, ..., f_m$ and their kernel sizes denote as $k_1, k_2, ..., k_m$. Therefore, $m$ 1-dimensional convolutions can be applied to the input matrix $E$. The convolutional procedure can be formalized as:

\begin{equation} \label{eq:filter}
  \begin{split}
  H_1= f_1(E)=\bigwedge \limits_{j=1}^{n} tanh(W^T_1 E^{j:j+k_1-1}),\\
  ...  \\
  H_m= f_m(E)=\bigwedge \limits_{j=1}^{n} tanh(W^T_m E^{j:j+k_m-1}),\\
  \end{split}
\end{equation}
where $\bigwedge \limits_{j=1}^{n}$ indicates the convolutional operations from left to right. Here we forced the row number $n$ of the output $H_1$ or $H_m \in \mathbb{R}^{n\times d^f}$ to be the same as that of the input $E$, because we aimed to keep the sequence length unchanged after convolution. It is simple to implement such goal, e.g., setting the kernel size, padding and stride as $k$, $floor(k/2)$ and 1. $d^f$ indicates the out-channel size of a filter and every filter has the same output size.

Moreover, $E^{j:j+k_1-1} \in \mathbb{R}^{k_1\times d^{e}}$ and $E^{j:j+k_m-1} \in \mathbb{R}^{k_m\times d^{e}}$ indicate the sub-matrices of $E$, starting from the $j$-th row and ending at the $j+k_1-1$ or $j+k_m-1$ row. $W_1 \in \mathbb{R}^{(k_1\times d^{e})\times d^f}$ and $W_m \in \mathbb{R}^{(k_m\times d^{e})\times d^f}$ indicate the weight matrices of corresponding filters. Throughout this paper, the biases of all layers are ignored for conciseness. The overview of a 1-dimensional convolution filter $f_m$ is shown in Figure \ref{fig:conv}.

\subsection{Residual Convolutional Layer}

\begin{figure}[t]
\centering
\includegraphics[width=.95\columnwidth]{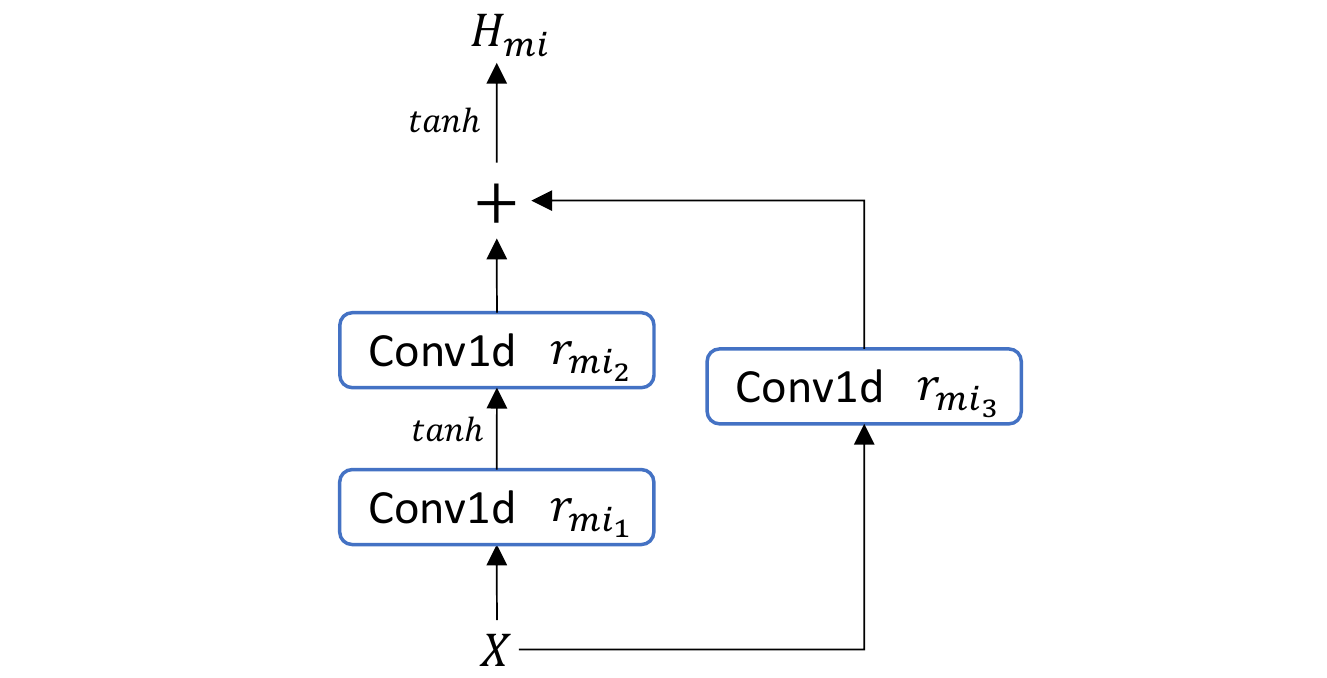} 
\caption{The architecture of a residual block $r_{mi}$. ``+'' represents the element-wise addition.}
\label{fig:resblock}
\end{figure}

On top of each filter in the multi-filter convolutional layer, there is a residual convolutional layer which consists of $p$ residual blocks \cite{he2016deep}. Take the $m$-th filter as an example, the computational procedure of its corresponding residual blocks $r_{m1}, r_{m2}, ..., r_{mp}$ can be formalized as:

\begin{algorithm}
% \small
\begin{algorithmic}[1]
		\STATE $X=H_m$
        \FOR{$i$ = 1 \TO $p$}
            \STATE  $H_{mi} = r_{mi}(X)$
            \STATE  $X=H_{mi}$
        \ENDFOR
        \RETURN $H_{mp}$
\end{algorithmic}
\end{algorithm}

For the residual block $r_{mi}$ (Figure \ref{fig:resblock}), it consists of three convolutional filters, namely $r_{{mi}_1}$, $r_{{mi}_2}$ and $r_{{mi}_3}$. The computational procedure can be denoted as:

\begin{equation} \label{eq:res}
  \begin{split}
  X_1= r_{{mi}_1}(X)=\bigwedge \limits_{j=1}^{n} tanh(W^T_{{mi}_1} X^{j:j+k_m-1}),\\
  X_2= r_{{mi}_2}(X_1)=\bigwedge \limits_{j=1}^{n} W^T_{{mi}_2} X_1^{j:j+k_m-1},\\
  X_3= r_{{mi}_3}(X)=\bigwedge \limits_{j=1}^{n} W^T_{{mi}_3} X^{j:j},\\
  H_{mi}=tanh(X_2+X_3),\\
  \end{split}
\end{equation}
where $\bigwedge \limits_{j=1}^{n}$ indicates the convolutional operations. $X$ denotes the input matrix of this residual block and $X^{j:j+k_m-1} \in \mathbb{R}^{k_m\times d^{i-1}}$ indicate the sub-matrices of $X$, starting from the $j$-th row and ending at the $j+k_m-1$ row. $H_{mi} \in \mathbb{R}^{n\times d^i}$ denotes the output matrix of the residual block. $d^{i-1}$ and $d^{i}$ denote the in-channel and out-channel sizes of this residual block. Therefore, the in-channel size of the first residual block $r_{m1}$ should be $d^f$ and the out-channel size of the last residual block $r_{mp}$ is defined as $d^p$. Similar with the multi-filter convolutional layer, we let the row numbers of $H_{mi}$ as well as $X_1$, $X_2$ and $X_3 \in \mathbb{R}^{n\times d^i}$ be $n$, which is identical to that of the input $X$.

Moreover, $W_{{mi}_1} \in \mathbb{R}^{(k_m\times d^{i-1})\times d^i}$, $W_{{mi}_2} \in \mathbb{R}^{(k_m\times d^i)\times d^i}$ and $W_{{mi}_3} \in \mathbb{R}^{(1\times d^{i-1})\times d^i}$ denote the weight matrices of the three convolutional filters, $r_{{mi}_1}$, $r_{{mi}_2}$ and $r_{{mi}_3}$. Thereinto, $r_{{mi}_1}$ and $r_{{mi}_2}$ have the same kernel size $k_m$ with the corresponding filter $f_m$ in the multi-filter convolutional layer, but they have different in-channel sizes. $r_{{mi}_3}$ is a special convolutional filter whose kernel size is 1.

Because the $m$-th filter $f_m$ in the multi-filter convolutional layer corresponds to $p$ residual blocks $r_{m1}, r_{m2}, ..., r_{mp}$ in the residual convolutional layer, we employed the output $H_{mp} \in \mathbb{R}^{n\times d^p}$ of the $p$-th residual block $r_{mp}$ as the output of these residual blocks. Since there are totally $m$ filters in the multi-filter convolutional layer, the final output of the residual convolutional layer is a concatenation of the output of $m$ residual blocks, namely $H=H_{1p}\oplus H_{2p}...H_{mp} \in \mathbb{R}^{n\times(m\times d^p)}$.

\subsection{Attention Layer}

Following Mullenbach et al. \shortcite{mullenbach2018explainable}, we employed the per-label attention mechanism to make each ICD code attend to different parts of the document representation $H$. The attention layer is formalized as:

\begin{equation} \label{eq:att}
  \begin{split}
  A =softmax(H U),\\
  V = A^T H,\\
  \end{split}
\end{equation}
where $U \in \mathbb{R}^{(m\times d^p)\times l}$ represents the parameter matrix of the attention layer, $A \in \mathbb{R}^{n\times l}$ represents the attention weights for each pair of an ICD code and a word, $V \in \mathbb{R}^{l\times(m\times d^p)}$ represents the output of the attention layer. Here $l$ denotes the number of ICD codes.

\subsection{Output Layer}
In the output layer, $V$ is first fed into a linear layer followed by the sum-pooling operation to obtain the score vector $\hat{y}$ for all ICD codes, and then the probability vector $\tilde{y}$ is calculated from $\hat{y}$ by the sigmoid function. This process can be formalized as:

\begin{equation} \label{eq:output}
  \begin{split}
  Y = V W, where\ Y \in \mathbb{R}^{l\times l},\\
  \hat{y} = pooling(Y), where\ \hat{y}_i = \sum \limits_{j=1}^{l} Y_{ij}, \\
  \tilde{y} =sigmoid(\hat{y}),\\
  \end{split}
\end{equation}
where $W \in \mathbb{R}^{(m\times d^p)\times l}$ is the weight matrix of the output layer. For training, we treated the ICD coding task as a multi-label classification problem following previous work \cite{mccallum1999multi,mullenbach2018explainable}. The training objective is to minimize the binary cross entropy loss between the prediction $\tilde{y}$ and the target $y$:

\begin{equation} \label{eq:loss}
  \begin{split}
  L(w,y,\theta) = -\sum \limits_{j=1}^{l} y_j log(\tilde{y}_j) + (1-y_j) log(1-\tilde{y}_j),
  \end{split}
\end{equation}
where $w$ denotes the input word sequence and $\theta$ denotes all the parameters. We utilized the back-propagation algorithm and Adam optimizer \cite{kingma2014adam} to train our model.

\begin{table*}[t]
\caption{Performance comparisons using different configurations in the multi-filter and residual convolutional layers. $k$ denotes the kernel sizes $k_1, k_2, ..., k_m$ and $p$ denotes the residual block number.}\smallskip
\centering
%\resizebox{0.95\textwidth}{!}{ % If your table exceeds the column or page width, use this command to reduce it slightly
\begin{tabular}{l|l|lll|lll}
\hline
\multirow{2}{*}{Model} & \multirow{2}{*}{Config} &  \multicolumn{3}{|c|}{MIMIC-III, full codes}  & \multicolumn{3}{|c}{MIMIC-III, top-50 codes}\\
 &  &  P@8  & Micro-F1 & Macro-F1 & P@5 & Micro-F1 & Macro-F1 \\
\hline
CNN & $k$=9 & 0.706  & 0.508 & 0.053 & 0.590 & 0.592 & 0.519\\
\hline
 \multirow{3}{*}{MultiCNN} & $k$=5,9,15 & 0.731  & 0.534 & 0.061 & 0.616 & 0.633 & 0.556\\
& $k$=3,5,9,15,19 &  0.735  & 0.542 & 0.067 & 0.630 & 0.646 & 0.576\\
& $k$=3,5,9,15,19,25 &  0.736  & 0.545 & 0.068 & 0.633& 0.652 & 0.584\\
\hline
\multirow{3}{*}{ResCNN}  & $p$=1 & 0.714 & 0.532  & 0.063 &  0.618 & 0.645 & 0.560\\
 & $p$=2 &  0.713 & 0.532  & 0.059 &0.589 & 0.601 & 0.531\\
 & $p$=3 &  0.710  & 0.529  & 0.059 & 0.575& 0.585 & 0.500\\
\hline
\multirow{2}{*}{MultiResCNN} & $k$=3,5,9,15,19,25 & \multirow{2}{*}{0.741} & \multirow{2}{*}{0.561}  & \multirow{2}{*}{0.073} & \multirow{2}{*}{0.638} & \multirow{2}{*}{0.673} & \multirow{2}{*}{0.608}\\
 & $p$=1 &  &  &  & \\
\hline
\end{tabular}
% }
\label{table:net_compare}
\end{table*}

\section{Experiments}
\subsection{Datasets}

\subsubsection{MIMIC-III}
In this paper, we employed the third version of \textbf{M}edical \textbf{I}nformation \textbf{M}art for \textbf{I}ntensive \textbf{C}are (MIMIC-III) \cite{johnson2016mimic} as the first dataset to evaluate our models. Following Mullenbach et al. \shortcite{mullenbach2018explainable}, we used discharge summaries, split them by patient IDs, and conducted experiments using the full codes as well as the top-50 most frequent codes. Finally, the MIMIC-III dataset using 8,921 ICD-9 codes consists of 47,719, 1,631 and 3,372 discharge summaries for training, development and testing respectively. The dataset using top-50 codes has 8,067 discharge summaries for training, 1,574 for development, and 1,730 for testing.

\subsubsection{MIMIC-II}
Besides the MIMIC-III dataset, we also leveraged the MIMIC-II dataset to compare our models with the ones in previous work \cite{perotte2013diagnosis,mullenbach2018explainable,baumel2018multi}. Follow their experimental setting, there are 20,533 and 2,282 clinical notes for training and testing, and 5,031 unique ICD-9 codes in the dataset.

\subsubsection{Preprocessing}
Following previous work \cite{mullenbach2018explainable}, the text was tokenized, and each token were transformed into its lowercase. The tokens that contain no alphabetic characters were removed such as numbers and punctuations. The maximum length of a token sequence is 2,500 and the one that exceeds this length will be truncated. We utilized the scripts\footnote{https://github.com/jamesmullenbach/caml-mimic} provided by Mullenbach et al. \shortcite{mullenbach2018explainable} for preprocessing.

\subsection{Evaluation Metrics}

To compare with previous work, we utilized different evaluation metrics in different experiments. In the MIMIC-III experiment using full ICD codes, we utilized macro-averaged and micro-averaged AUC (area under the ROC, i.e., receiver operating characteristic curve), macro-averaged and micro-averaged F1, precision at 8 (P@8) and precision at 15 (P@15). When computing macro-averaged AUC or F1, we first computed the performance for each label and then averaged them. When computing micro-averaged AUC or F1, we considered every pair of a clinical note and a code as an independent prediction. The precision at K (P@K) indicates the proportion of the correctly-predicted labels in the top-K predicted labels. 

In the MIMIC-III experiment using the top-50 ICD codes, we employed the P@5 besides macro-averaged and micro-averaged AUC, macro-averaged and micro-averaged F1. In the MIMIC-II experiment using full codes, we employed the same evaluation metrics except that P@5 was changed to P@8.

\begin{table*}[t]
\centering
\caption{MIMIC-III results (full codes). The results of MultiResCNN are shown in means $\pm$ standard deviations.}\smallskip
%\resizebox{0.95\textwidth}{!}{ % If your table exceeds the column or page width, use this command to reduce it slightly
\begin{tabular}{l|ll|ll|ll}
\hline
& \multicolumn{2}{|c|}{AUC} & \multicolumn{2}{|c|}{F1} & \multicolumn{2}{|c}{P@K} \\
Model & Macro & Micro & Macro & Micro &  8 & 15 \\
\hline
CAML \cite{mullenbach2018explainable} & 0.895 & \textbf{0.986} & \textbf{0.088} & 0.539 &  0.709 & 0.561 \\
DR-CAML \cite{mullenbach2018explainable} & 0.897 & 0.985 & 0.086 & 0.529 & 0.690 & 0.548 \\
\hline
\multirow{2}{*}{MultiResCNN}  & \textbf{0.910} & \textbf{0.986} & 0.085 & \textbf{0.552} & \textbf{0.734} & \textbf{0.584} \\
 & $\pm$0.002 & $\pm$0.001 & $\pm$0.007 & $\pm$0.005 & $\pm$0.002 & $\pm$0.001 \\
\hline
\end{tabular}
%}
\label{table:mimic3full}
\end{table*}

\begin{table*}[t]
\centering
\caption{MIMIC-III results (top-50 codes). The results of MultiResCNN are shown in means $\pm$ standard deviations.}\smallskip
%\resizebox{0.95\textwidth}{!}{ % If your table exceeds the column or page width, use this command to reduce it slightly
\begin{tabular}{l|ll|ll|l}
\hline
& \multicolumn{2}{|c|}{AUC} & \multicolumn{2}{|c|}{F1} &  \\
Model & Macro & Micro & Macro & Micro & P@5 \\
\hline
C-MemNN \cite{prakash2017condensed} & 0.833 & - & - & - & 0.420 \\
C-LSTM-Att \cite{shi2017towards} & - & 0.900 & - & 0.532 & - \\
CAML \cite{mullenbach2018explainable} & 0.875 & 0.909 & 0.532 & 0.614 & 0.609  \\
DR-CAML \cite{mullenbach2018explainable} & 0.884 & 0.916 & 0.576 & 0.633 & 0.618  \\
\hline
\multirow{2}{*}{MultiResCNN}  & \textbf{0.899}  & \textbf{0.928} & \textbf{0.606} & \textbf{0.670} &  \textbf{0.641} \\
& $\pm$0.004  & $\pm$0.002 & $\pm$0.011 & $\pm$0.003 &  $\pm$0.001 \\
\hline
\end{tabular}
%}
\label{table:mimic3_50}
\end{table*}

\subsection{Hyper-parameter Tuning}
Since our model has a number of hyper-parameters, it is infeasible to search optimal values for all hyper-parameters. Therefore, some hyper-parameter values were chosen empirically or following prior work \cite{mullenbach2018explainable}. The word embedding size $d^e$ is 100, the out-channel size $d^f$ of a filter in the multi-filter convolutional layer is 100, the learning rate is 0.0001, the batch size is 16 and the dropout rate is 0.2. 

To explore a better configuration for the filter number $m$ and the kernel sizes $k_1, k_2, ..., k_m$ in the multi-filter convolutional layer, and the residual block number $p$ in the residual convolutional layer, we conducted the following experiments. First, we developed three variations:

\begin{itemize}
\item CNN, which only has one convolutional filter and is equivalent to the CAML model \cite{mullenbach2018explainable}.
\item MultiCNN, which only has the multi-filter convolutional layer.
\item ResCNN, which only has the residual convolutional layer.
\end{itemize}

Then we tried several configurations for these models on the development set of MIMIC-III using the full and top-50 code settings. The experimental results are shown in Table \ref{table:net_compare}. For each configuration, we tried three runs by initializing the model parameters randomly. The results shown in the table are the means of three runs. We selected such kernel sizes since they do not only capture various text patterns from different granularities, but also
keeps the sequence length unchanged after convolution (e.g., setting the padding and stride sizes as floor(k/2) and 1). In addition, we pre-defined the in-channel and out-channel sizes of residual blocks empirically:

\begin{itemize}
\item $p$=1: $d^0$=100, $d^1$=50
\item $p$=2: $d^0$=100, $d^1$=100, $d^2$=50
\item $p$=3: $d^0$=100, $d^1$=150, $d^2$=100, $d^3$=50
\end{itemize}

As shown in Table \ref{table:net_compare}, MultiCNN performs better than CNN. As the kernel number increases, the performance increases consistently in both full and top-50 code settings. The performance reaches a peak when the kernel sizes are 3,5,9,15,19,25. Moreover, ResCNN also performs better than CNN, but the difference is that the performances deteriorate as the residual block number increases. ResCNN achieves the best performance when the residual block number is 1. Therefore, we applied the best configuration of MultiCNN and ResCNN to MultiResCNN. The results show that the performance of MultiResCNN was further improved after combining MultiCNN and ResCNN. Therefore, we kept such configuration in other experiments.

\subsection{Baselines}

\subsubsection{CAML \& DR-CAML}
The \textbf{C}onvolutional \textbf{A}ttention network for \textbf{M}ulti-\textbf{L}abel classification (CAML) was proposed by Mullenbach et al. \shortcite{mullenbach2018explainable}. It has achieved the state-of-the-art results on the MIMIC-III and MIMIC-II datasets among the models using unstructured text. It consists of one convolutional layer and one attention layer to generate label-aware features for multi-label classification \cite{mccallum1999multi}. The \textbf{D}escription \textbf{R}egularized CAML (DR-CAML) is an extension of CAML and incorporates the text description of each code to regularize the model.

\subsubsection{C-MemNN}
The \textbf{C}ondensed \textbf{Mem}ory \textbf{N}eural \textbf{N}etwork was proposed by Prakash et al. \shortcite{prakash2017condensed}, which equips the neural network with iterative condensed memory representations. The model achieved competitive results to predict the top-50 ICD codes for the medical notes in the MIMIC-III dataset.

\subsubsection{C-LSTM-Att}
Shi et al. \shortcite{shi2017towards} proposed a \textbf{C}haracter-aware \textbf{LSTM}-based \textbf{A}ttention model to assign ICD codes to clinical notes. They employed LSTM-based language models to generate representations of clinical notes and ICD codes, and proposed an attention method to address the mismatch between notes and codes. They also focused on predicting the top-50 ICD codes for the medical notes in the MIMIC-III dataset.

\subsubsection{SVM}
Perotte et al. \shortcite{perotte2013diagnosis} experimented two approaches: one treats each ICD9 code independently (flat SVM) and the other uses the hierarchical nature of ICD9 codes (hierarchy SVM). Their results show that the hierarchy SVM performs better than the flat one, yielding 29.3\% f1-measure in the MIMIC-II dataset.

\subsubsection{HA-GRU}
Baumel et al. \shortcite{baumel2018multi} presented a model named \textbf{H}ierarchical \textbf{A}ttention \textbf{G}ated \textbf{R}ecurrent \textbf{U}nit (HA-GRU) for automatic ICD coding of clinical documents. HA-GRU includes two main layers: the first one encodes sentences and the second one encodes documents. They reported their results in the MIMIC-II dataset, following the data split from Perotte et al. \shortcite{perotte2013diagnosis}.

\subsection{Results}
In this section, we compared our model with existing work for automated ICD coding. We ran our model three times for each experiment and each time we used different random seeds for parameter initialization. The final results are the means and standard deviations of three runs. Following prior work \cite{mullenbach2018explainable}, we compared our model with existing work using the MIMIC-III and MIMIC-II dataset. For the MIMIC-III dataset, we also performed the comparisons with two experimental settings, namely using the full codes and top-50 codes. For the MIMIC-II dataset, only the full codes were employed.

\subsubsection{MIMIC-III Results (full codes)}

As shown in Table \ref{table:mimic3full}, we can see that our model obtained better results in the macro-AUC, micro-F1, precision@8 and precision@15, compared with the state-of-the-art models, CAML and DR-CAML. Our model improved the macro-AUC by 0.013, the micro-F1 by 0.013, the precision@8 by 0.025, the precision@15 by 0.023. In addition, our model achieved comparable performance on the micro-AUC and a slightly worse macro-F1. More importantly, we observed that our model is able to attain stable good results from the standard deviations.

\begin{table*}[t]
\centering
\caption{MIMIC-II results (full codes). The results of MultiResCNN are shown in means $\pm$ standard deviations.}\smallskip
%\resizebox{0.95\textwidth}{!}{ % If your table exceeds the column or page width, use this command to reduce it slightly
\begin{tabular}{l|ll|ll|l}
\hline
& \multicolumn{2}{|c|}{AUC} & \multicolumn{2}{|c|}{F1} &  \\
Model & Macro & Micro & Macro & Micro & P@8 \\
\hline
SVM \cite{perotte2013diagnosis} & - & - & - & 0.293 & - \\
HA-GRU \cite{baumel2018multi} & - & - & - & 0.366 & - \\ 
CAML \cite{mullenbach2018explainable} & 0.820 & 0.966 & 0.048 & 0.442 & 0.523  \\
DR-CAML \cite{mullenbach2018explainable} & 0.826 & 0.966 & 0.049 & 0.457 & 0.515  \\
\hline
\multirow{2}{*}{MultiResCNN} & \textbf{0.850} & \textbf{0.968} & \textbf{0.052} & \textbf{0.464} &  \textbf{0.544} \\
& $\pm$0.002  & $\pm$0.001 & $\pm$0.002 & $\pm$0.002 &  $\pm$0.007 \\
\hline
\end{tabular}
%}
\label{table:mimic2_full}
\end{table*}

\subsubsection{MIMIC-III Results (top-50 codes)}

From Table \ref{table:mimic3_50}, we observed that our model outperformed all the baselines, namely C-MemNN \cite{prakash2017condensed}, C-LSTM-Att \cite{shi2017towards}, CAML and DR-CAML \cite{mullenbach2018explainable}, in all evaluation metrics. Our model improves the macro-AUC, micro-AUC, macro-F1, micro-F1 and precision@5 by 0.015, 0.012, 0.030, 0.037 and 0.023, respectively. Our model outperformed the C-MemNN by 0.221 and 0.066 in precision@5 and macro-AUC. It also outperformed the C-LSTM-Att by 0.138 and 0.028 in micro-F1 and micro-AUC. Its precision@5 is 0.032 and 0.023 higher than those of CAML and DR-CAML.

\subsubsection{MIMIC-II Results (full codes)}

Table \ref{table:mimic2_full} shows the results on the full code set of MIMIC-II. Perotte et al. \shortcite{perotte2013diagnosis} used the SVM to predict ICD codes from clinical text and their method obtained 0.293 micro-F1. By contrast, our model outperformed their method by 0.171 in micro-F1. Baumel et al. \shortcite{baumel2018multi} utilized the attention mechanism and GRU for automated ICD coding. Our model outperformed their model by 0.098 in micro-F1. Our model also outperformed the state-of-the-art model, CAML or DR-CAML, by 0.024, 0.002, 0.003, 0.007 and 0.021 in all evaluation metrics.

\section{Discussion}
\subsection{Computational Cost Analysis}
In this section, we analyzed the computational cost between the state-of-the-art model, CAML and our model, MultiResCNN. The analysis was conducted from four aspects, namely the parameter amount, training time, training epoch, inference speed. Our experimental settings are as follows. For CAML, we used the optimal hyper-parameter setting reported in their paper \cite{mullenbach2018explainable}. For MultiResCNN, we used six filters and 1 residual block, which obtained the best result in our hyper-parameter tuning experiments. The batch size, learning rate and dropout rate are identical in every experiment. We used the training set and development set of MIMIC-III (full codes) as experimental data. The experiments were conducted on NVIDIA Tesla P40 GPUs. Training will terminate if the performance on the development set does not increase for 10 times.

\begin{table}[t]
\caption{Analysis of the computational cost between CAML and MultiResCNN. ``m'', ``s'', ``ep'' and ``d'' denote million, second, epoch and document respectively.}\smallskip
\centering
% \resizebox{.95\columnwidth}{!}{
% \smallskip
\begin{tabular}{l|l|l}
\hline
 & CAML & MultiResCNN\\
\hline
Parameter Amount & 6.2m & 11.9m \\
Training Time & 438s/ep  & 1026s/ep  \\
Training Epoch & 85 & 26 \\
Inference Speed & 108.7d/s & 70.9d/s \\
\hline
\end{tabular}
% }
\label{table:cost}
\end{table}

As shown in Table \ref{table:cost}, the parameter of MultiResCNN is approximately 1.9 times as many as that of CAML. The training time of MultiResCNN is about 2.3 times more than that of CAML. It is reasonable since MultiResCNN has more filters and layers. Interestingly, MultiResCNN needs much less epochs to converge. Considering the inference speed, CAML is approximately 1.5 times faster than MultiResCNN. Overall, the computational cost of MultiResCNN is larger than that of CAML, but we hold the opinion that the increased cost is still acceptable.

\subsection{Effect of Truncating Data}

During preprocessing, we truncated the discharge summaries that are longer than 2,500 tokens. To investigate the effect of the length limitation, we further conducted the experiments using 3,500, 4,500, 5,500 and 6,500. We selected these values because the maximum length of the discharge summaries in the development set is approximately 6,300. Results show that the performance differences between different settings are not significant. P@8 ranges between 0.736 and 0.741, and micro-F1 ranges between 0.557 and 0.566. 2,500 seems to be a decent selection considering the tradeoff between performance and cost.

\subsection{Limitations}
In this study, the performance improvement mostly comes from deep and diversified representations of text. In the future, we will explore how to incorporate BERT \cite{devlin-etal-2019-bert} into this task effectively and efficiently. In our preliminary experiments, BERT did not perform well due to the limitations of hardware and its fixed-length context. Therefore, potential solutions include recurrent Transformer \cite{dai-etal-2019-transformer} and hierarchical BERT \cite{zhang-etal-2019-hibert}. Moreover, we chose the kernel sizes of the multi-filter layer and channel sizes of the residual layer empirically, which should be further studied and optimized in the future.

\section{Conclusions}

In this paper, we proposed a multi-filter residual convolutional neural network for ICD coding. We conducted three experiments on the widely-used MIMIC-III and MIMIC-II datasets. Results show that our model achieved the state-of-the-art performance compared with several competitive baselines. We found that both multi-filter convolution and residual convolution helped the performance improvement with acceptable computational cost. This shows deep and diversified text representations could benefit the ICD coding from clinical text. Our model can be a strong baseline for not only ICD coding, but also other text classification tasks. 

\section{Acknowledgments}
This work was supported in part by the Center for Intelligent Information Retrieval, R01DA045816, R01HL125089, R01HL137794, R01HL135219, and R01LM012817. Any opinions, findings and conclusions or recommendations expressed in this material are those of the authors and do not necessarily reflect those of the sponsor.

\bibliography{icd_coding}
\bibliographystyle{aaai}
\end{document}